\documentclass[11pt]{article}

\usepackage{acl}
\usepackage{amsmath}
\usepackage{times}
\usepackage{latexsym}
\usepackage{booktabs}
\usepackage[table]{xcolor}
\usepackage{multirow}
\usepackage{amsfonts}
\usepackage{array} 

\usepackage[T1]{fontenc}

\usepackage[utf8]{inputenc}

\usepackage{microtype}

\usepackage{inconsolata}

\usepackage{graphicx}
\usepackage[edges]{forest}
\usepackage{tikz}
\usetikzlibrary{trees}

\tikzset{
    parent/.style={align=left,rounded corners=3pt},
    child/.style={align=left,rounded corners=3pt}
}
\title{Large Language Models Meet Virtual Cell: A Survey}

\author{
  \textbf{Krinos Li\textsuperscript{1}\thanks{Equal contribution.}},
  \textbf{Xianglu Xiao\textsuperscript{1}\footnotemark[1]},
  \textbf{Shenglong Deng\textsuperscript{1}\footnotemark[1]},
  \textbf{Lucas He\textsuperscript{2}\footnotemark[1]},
  \textbf{Zijun Zhong\footnotemark[1]},\\
  \textbf{Yuanjie Zou\textsuperscript{3}},
  \textbf{Zhonghao Zhan\textsuperscript{1}},
  \textbf{Zheng Hui\textsuperscript{4}},
 \textbf{Weiye Bao\textsuperscript{1}},
  \textbf{Guang Yang\textsuperscript{1,5,6}},
\\
  \textsuperscript{1}Imperial College London,
\textsuperscript{2}University College London,
  \textsuperscript{3}New Jersey Institute of Technology,\\
\textsuperscript{4}University of Cambridge,
  \textsuperscript{5}King's College London,
  \textsuperscript{6}Royal Brompton Hospital
}

\begin{document}
\maketitle
\begin{abstract}
Large language models (LLMs) are transforming cellular biology by enabling the development of "virtual cells"—computational systems that represent, predict, and reason about cellular states and behaviors. This work provides a comprehensive review of LLMs for virtual cell modeling. We propose a unified taxonomy that organizes existing methods into two paradigms: LLMs as Oracles, for direct cellular modeling, and LLMs as Agents, for orchestrating complex scientific tasks. We identify three core tasks—cellular representation, perturbation prediction, and gene regulation inference—and review their associated models, datasets, evaluation benchmarks, as well as critical challenges in scalability, generalizability, and interpretability.
\end{abstract}

\section{Introduction}
\begin{figure*}[!ht]
  \centering
  \includegraphics[width=0.9\textwidth]{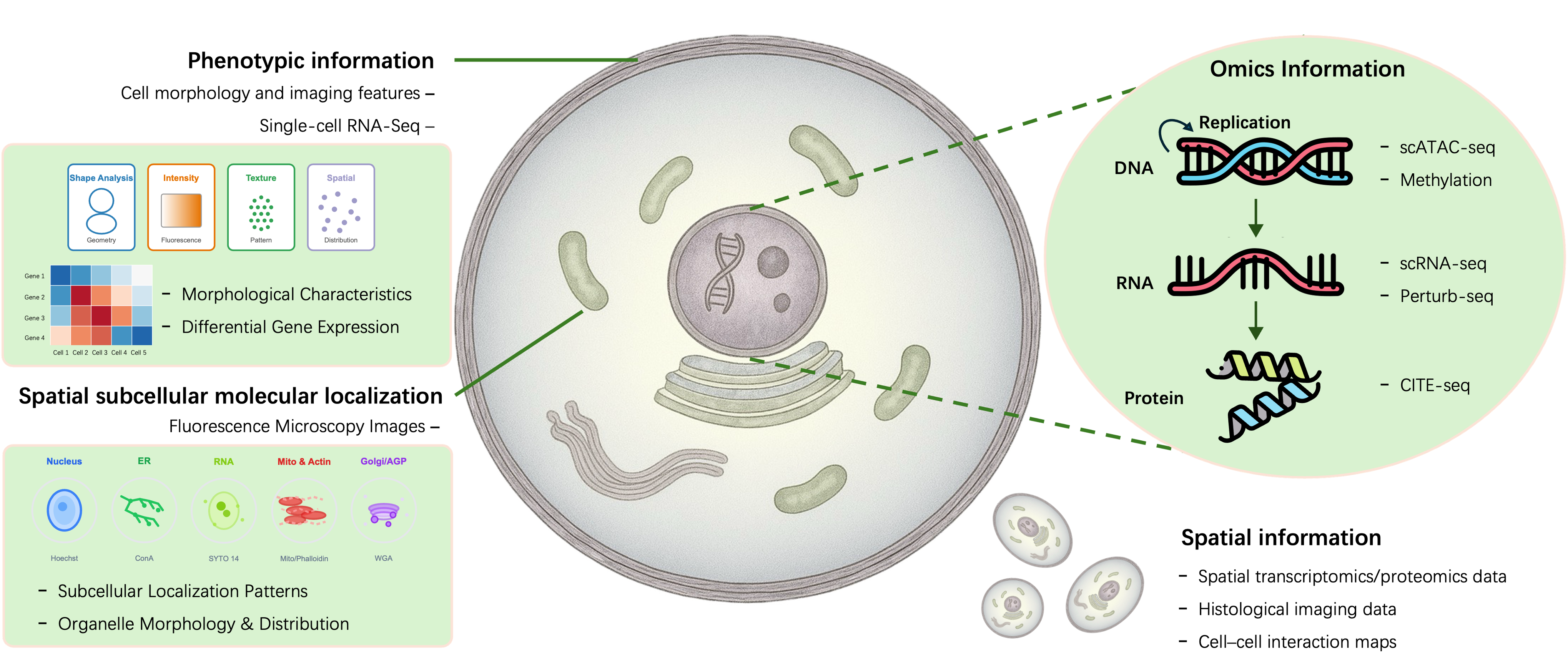}
  \caption{An illustration of the cell’s multiscale organization.}
  \label{fig:cell}
\end{figure*}

Cells are the fundamental units of life that execute intricate molecular programs that drive proliferation, differentiation, and homeostasis \citep{polychronidouSinglecellBiologyWhat2023}. Understanding how these programs give rise to cellular behavior has long been a central goal of biology, yet the enormous complexity and high dimensionality of molecular interactions have made this task daunting (Fig. \ref{fig:cell}). Recent advances in artificial intelligence (AI), particularly large language models (LLMs), have opened an unprecedented opportunity to bridge this gap by enabling the concept of a \textit{virtual cell}: a computational system that emulates the structure, function and dynamics of cellular cells in silico \citep{szalata2024transformers,cui2025towards}. Such systems have transformative potential, from accelerating drug discovery to enabling personalized medicine through predictive cellular models \citep{bunne2024build}.

The notion of a virtual cell is not entirely new; early systems biology sought to reconstruct cellular behavior through mechanistic or statistical modeling\citep{qiao2024evolution,schmidt2013mechanistic}. However, these approaches were limited by incomplete knowledge and data sparsity\citep{schmidt2013mechanistic}. With the explosion of omics data and the rise of LLMs, researchers can now train foundation models directly on large-scale biological corpora—ranging from nucleotide sequences and single-cell transcriptomes to multi-omic and spatial data—allowing the virtual cell to emerge as a data-driven, generative, and reasoning framework\citep{szalata2024transformers,cui2025towards}. 

The growing availability of comprehensive datasets and large-scale research programs has further accelerated this trend. For example, the Joint Undertaking for Morphological Profiling (JUMP-Cell Painting) consortium has released standardized, multimodal datasets that provide rich resources for virtual cell model development and validation\citep{chandrasekaran2024three}. Similarly, the Chan Zuckerberg Initiative (CZI) has invested heavily in building open resources such as CELLxGENE and the Tabula Sapiens project, catalyzing collaborative data sharing across the scientific community\citep{thomas2025release}. Combined with the rapid rise of AI-powered single-cell studies and foundation model research, these collective efforts have positioned the virtual cell as one of the most rapidly advancing and influential frontiers in modern computational biology.

\begin{figure*}[t]
  \centering
  \includegraphics[width=0.9\textwidth]{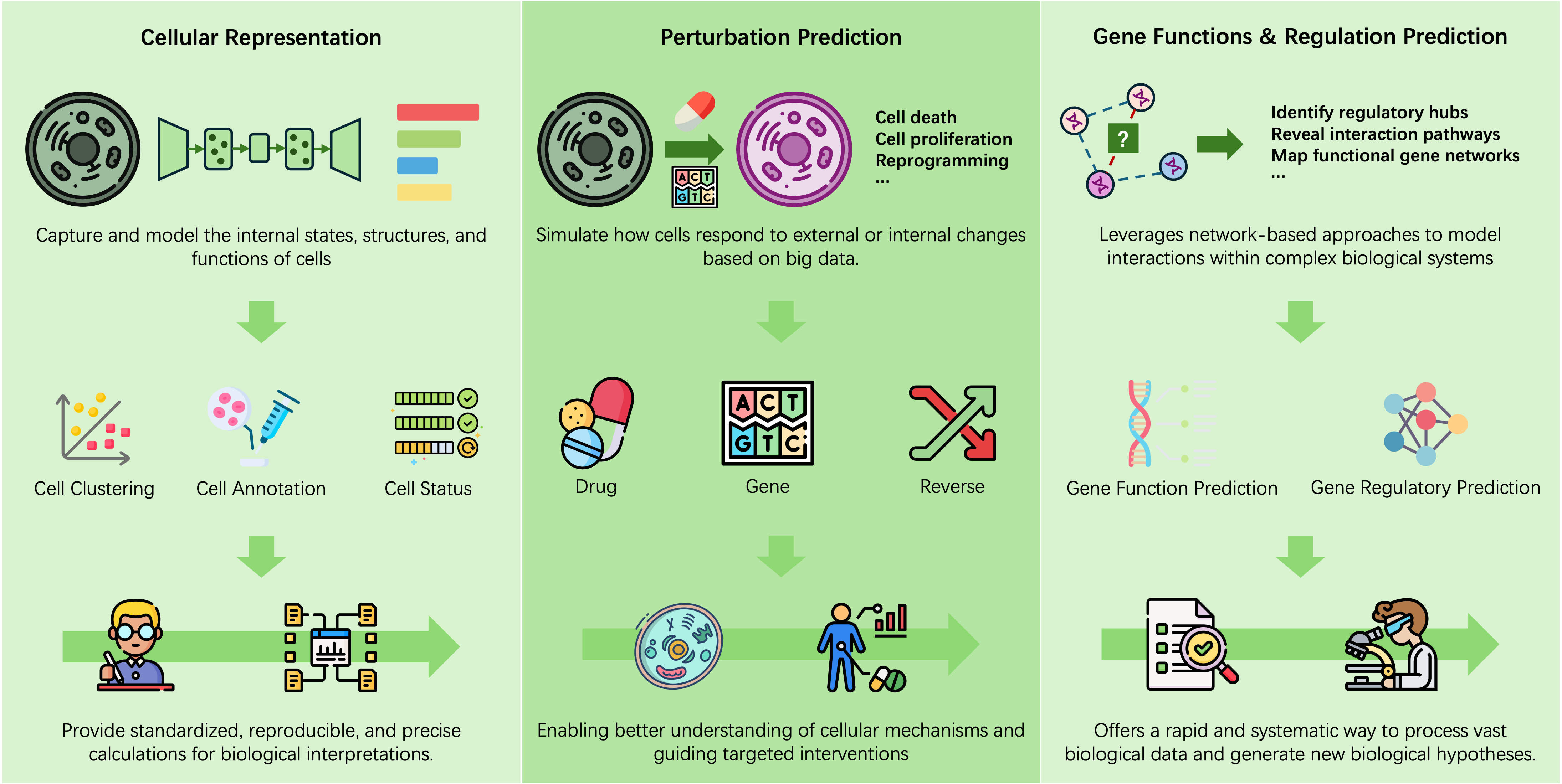}
  \caption{An overview of major tasks in AI-based virtual cell modeling}
  \label{fig:problems}
\end{figure*}

These advances have collectively established a new foundation for modeling cellular systems with unprecedented scope and precision. Central to this endeavor are three core tasks (Fig. \ref{fig:problems}): (1) \textit{Cellular Representation}, which enables accurate cell annotation, classification, and state prediction essential for cellular status interpretation; (2) \textit{Perturbation Prediction}, which models the effects of genetic or drug interventions (and their inverses) to support causal inference and therapeutic discovery; and (3) \textit{Gene Function \& Regulation Prediction}, which deciphers gene roles and reconstructs regulatory networks to uncover the mechanistic logic underlying cellular processes. Together, these tasks define the operational pillars of an AI-driven virtual cell.

This review provides a comprehensive synthesis of how LLMs are redefining the concept of the virtual cell. The main contribution summarized as follows:
\begin{itemize}
    \item \textbf{Comprehensive Survey.} To the best of our knowledge, this is the first review to systematically summarize how LLMs and agents are transforming the development of the virtual cell, bridging artificial intelligence and cellular biology.  
    \item \textbf{Unified Framework.} We propose a coherent taxonomy that organizes existing methods into two complementary paradigms: \textit{LLMs as Oracles} for modeling cellular states and molecular interactions, and \textit{LLMs as Agents} for autonomous reasoning, planning, and experimentation, along with associated datasets, benchmarks, and evaluation protocols.  
    \item \textbf{Future Outlook.} By integrating current progress and identifying open challenges in scalability, interpretability, and biological fidelity, this review provides strategic insights and a roadmap for advancing next-generation AI-powered virtual cell systems.  
\end{itemize}

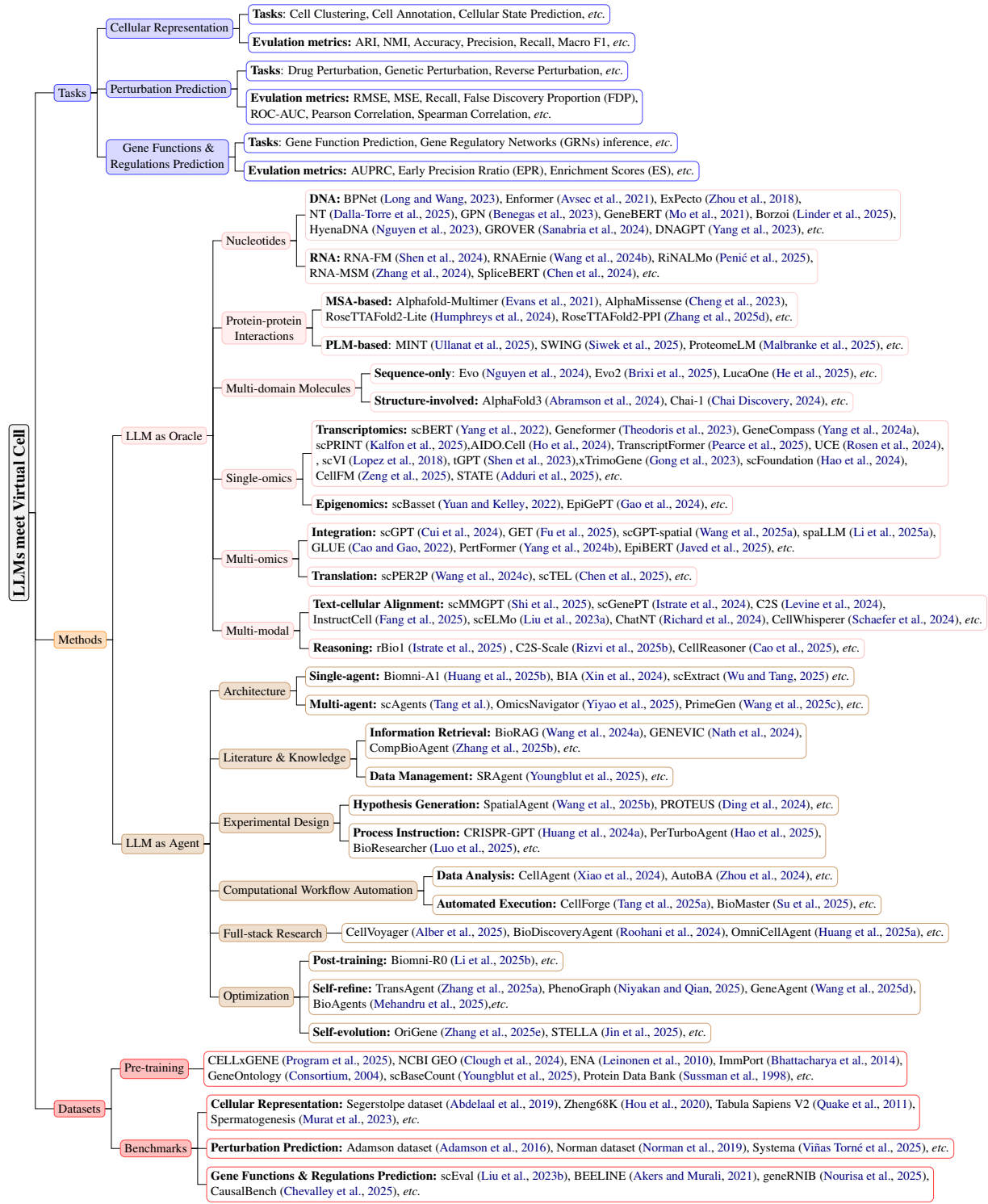
\begin{figure*}[!ht]
\resizebox{1\textwidth}{!}{
\begin{forest}
    for tree={
        grow'=east,
        forked edges,
        draw,
        rounded corners,
        node options={align=left},
        anchor=west
    }
    [LLMs meet Virtual Cell, rotate=90, font=\Large\bfseries, fill=gray!10, parent
        [Tasks, fill=blue!15,draw=blue!75, child,
            for children={fill=blue!15,draw=blue!75}
            [Cellular Representation, for children={draw=blue!75}
                [{\textbf{Tasks}: Cell Clustering, Cell Annotation, Cellular State Prediction, \textit{etc.}}]
                [{\textbf{Evulation metrics:} ARI, NMI,  Accuracy, Precision, Recall, Macro F1, \textit{etc.}}]
            ]
            [Perturbation Prediction, for children={draw=blue!75}
                [ {\textbf{Tasks}: Drug Perturbation, Genetic Perturbation, Reverse Perturbation, \textit{etc.}}]
                [{\textbf{Evulation metrics:} RMSE, MSE, Recall, False Discovery Proportion (FDP), \\ROC-AUC, Pearson Correlation, Spearman Correlation, \textit{etc.}}]
            ]
            [\quad Gene Functions \& \\Regulations Prediction, for children={draw=blue!75}
                [ {\textbf{Tasks}: Gene Function Prediction, Gene Regulatory Networks (GRNs) inference, \textit{etc.}}]
                [{\textbf{Evulation metrics:} AUPRC, Early Precision Rratio (EPR), Enrichment Scores (ES), \textit{etc.}}]
            ]
        ]
        [Methods, fill=orange!25,draw=orange!75, child,
            for children={fill=orange!25,draw=orange!75}
            [LLM as Oracle, fill=pink!25,draw=pink!95, child,for children={fill=pink!25,draw=pink!95}
            [Nucleotides, for children={draw=pink!75} [{\textbf{DNA:} BPNet \citep{long2023bpnet}, Enformer \citep{avsec2021effective}, ExPecto \citep{zhou2018deep}, \\NT \citep{dalla2025nucleotide},  GPN \citep{benegas2023dna}, GeneBERT \citep{mo2021multi}, Borzoi \citep{linder2025predicting}, \\HyenaDNA \citep{nguyen2023hyenadna}, GROVER \citep{sanabria2024dna}, DNAGPT \citep{yang2023dna}, \textit{etc.}}]
                [{\textbf{RNA:} RNA-FM \citep{shen2024accurate}, RNAErnie \citep{wang2024multi}, RiNALMo \citep{penic2025rinalmo}, \\RNA-MSM \citep{zhang2024multiple}, SpliceBERT \citep{chen2024self}, \textit{etc.}}]
                ]
            [Protein-protein\\\; \,Interactions, for children={draw=pink!75}
                [{\textbf{MSA-based:} Alphafold-Multimer \citep{evans2021protein}, AlphaMissense \citep{cheng2023accurate}, \\RoseTTAFold2-Lite \citep{humphreys2024protein}, RoseTTAFold2-PPI \citep{zhang2025predicting}, \textit{etc.}} ]
                [{\textbf{PLM-based}: MINT \citep{ullanat2025learning}, SWING \citep{siwek2025sliding}, ProteomeLM \citep{malbranke2025proteomelm}, \textit{etc.}}]
            ]
            [Multi-domain Molecules, for children={draw=pink!75}
                [{\textbf{Sequence-only}: Evo \citep{nguyen2024sequence}, Evo2 \citep{brixi2025genome}, LucaOne \citep{he2025generalized}, \textit{etc.}}]
                [{\textbf{Structure-involved:} AlphaFold3 \citep{abramson2024accurate}, Chai-1 \citep{Chai-1-Technical-Report}, \textit{etc.}}]
            ]
            [Single-omics, for children={draw=pink!95}
            [{\textbf{Transcriptomics:} scBERT \citep{yang2022scbert}, Geneformer \citep{theodoris2023transfer},  GeneCompass \citep{yang2024genecompass}, \\scPRINT \citep{kalfon2025scprint},AIDO.Cell \citep{ho2024scaling}, TranscriptFormer \citep{pearce2025cross}, UCE \citep{rosen2024toward}, \\, scVI \citep{lopez2018deep}, tGPT \cite{shen2023generative},xTrimoGene \citep{gong2023xtrimogene}, scFoundation \citep{hao2024large}, \\CellFM \citep{zeng2025cellfm}, STATE \citep{adduri2025predicting}, \textit{etc.}}]
            [{\textbf{Epigenomics:} scBasset \citep{yuan2022scbasset}, EpiGePT \citep{gao2024epigept},  \textit{etc.}}]
            ]
            [Multi-omics, for children={draw=pink!95}
            [{\textbf{Integration:} scGPT \citep{cui2024scgpt}, GET \citep{fu2025foundation}, scGPT-spatial \citep{wang2025scgpt}, spaLLM \citep{li2025spallm}, \\GLUE \citep{cao2022multi}, PertFormer \citep{yang2024multiomic}, EpiBERT \citep{javed2025multi}, \textit{etc.}}]
            [{\textbf{Translation:} scPER2P \citep{wang2024scper2p}, scTEL \citep{chen2025joint}, \textit{etc.}}]
            ]
            [Multi-modal, for children={draw=pink!95}
            [{\textbf{Text-cellular Alignment:} scMMGPT \citep{shi2025language}, scGenePT \citep{istrate2024scgenept}, C2S \citep{levine2024cell2sentence}, \\InstructCell \citep{fang2025multi}, scELMo \citep{liu2023scelmo}, ChatNT \citep{richard2024chatnt}, CellWhisperer \citep{schaefer2024multimodal}, \textit{etc.}}]
            [{\textbf{Reasoning:} rBio1 \citep{istrate2025rbio1}
            , C2S-Scale \citep{rizvi2025scaling}, CellReasoner \cite{cao2025cellreasoner}, \textit{etc.}}]
            ]
        ]     
        [LLM as Agent, fill=brown!25,draw=brown!75, child,for children={fill=brown!25,draw=brown!75}
            [Architecture, for children={draw=brown!75}
                [{\textbf{Single-agent:} Biomni-A1 \citep{huang2025biomni}, BIA \citep{xin2024bioinformatics}, scExtract \citep{wu2025scextract} 
                \textit{etc.}}]
                [{\textbf{Multi-agent:} scAgents \citep{tang2025scagents}, OmicsNavigator \citep{yiyao2025omicsnavigator}, PrimeGen \citep{wang2025accelerating}, \textit{etc.}}]
            ]
            [Literature \& Knowledge, for children={draw=brown!75}
            [{\textbf{Information Retrieval:} BioRAG \citep{wang2024biorag},  GENEVIC \citep{nath2024genevic}, \\CompBioAgent \citep{zhang2025compbioagent}, \textit{etc.}}]
            [{\textbf{Data Management:} SRAgent \citep{youngblut2025scbasecount}, \textit{etc.}}]
            ]
            [Experimental Design, for children={draw=brown!75}
                [{\textbf{Hypothesis Generation:}  SpatialAgent \citep{wang2025spatialagent}, PROTEUS \citep{ding2024automating}, \textit{etc.}}]
                [{\textbf{Process Instruction:} CRISPR-GPT \citep{huang2024crispr}, PerTurboAgent \citep{hao2025perturboagent}, \\BioResearcher \citep{luo2025intention}, \textit{etc.}}]
            ]
            [Computational Workflow Automation, for children={draw=brown!75}
                [{\textbf{Data Analysis:} CellAgent \citep{xiao2024cellagent}, AutoBA \citep{zhou2024ai}, \textit{etc.}}]
                [{\textbf{Automated Execution:} CellForge \citep{tang2025cellforge}, BioMaster \citep{su2025biomaster}, \textit{etc.}}]
            ]
            [Full-stack Research, for children={draw=brown!75}
                [{CellVoyager \citep{alber2025cellvoyager}, BioDiscoveryAgent \citep{roohani2024biodiscoveryagent}, OmniCellAgent \citep{huang2025omnicellagent}, \textit{etc.}
                }]
            ]
            [Optimization, for children={draw=brown!75}
                [{\textbf{Post-training:} Biomni-R0 \citep{biomnir0}, \textit{etc.}}]
                [{\textbf{Self-refine:} TransAgent \citep{zhang2025transagent}, PhenoGraph \citep{niyakan2025phenograph}, GeneAgent \citep{wang2025geneagent}, \\BioAgents \citep{mehandru2025bioagents},\textit{etc.}}]
                [{\textbf{Self-evolution:} OriGene \citep{zhang2025origene}, STELLA \citep{jin2025stella}, \textit{etc.}}]
            ]
        ] 
        ]
        [Datasets,  fill=red!25,draw=red!75, child,for children={fill=red!25,draw=red!75}
            [Pre-training, for children={draw=red!75}
                [{CELLxGENE \citep{czi2025cz}, NCBI GEO \citep{clough2024ncbi}, ENA \citep{leinonen2010european}, ImmPort \citep{bhattacharya2014immport}, \\GeneOntology \citep{gene2004gene}, scBaseCount \citep{youngblut2025scbasecount}, Protein Data Bank \citep{sussman1998protein}, \textit{etc.}}]
            ]
            [Benchmarks, for children={draw=red!75}
                [{\textbf{Cellular Representation:} Segerstolpe dataset \citep{abdelaal2019comparison}, Zheng68K \citep{hou2020systematic}, Tabula Sapiens V2 \citep{quake2011tabula}, \\Spermatogenesis \citep{murat2023molecular}, \textit{etc.}}]
                [{\textbf{Perturbation Prediction:} Adamson dataset \citep{adamson2016multiplexed}, Norman dataset \citep{norman2019exploring}, Systema \citep{vinas2025systema}, \textit{etc.}}]
                [{\textbf{Gene Functions \& Regulations Prediction:} scEval \citep{liu2023evaluating}, BEELINE \citep{akers2021gene}, geneRNIB \citep{nourisa2025genernib}, \\CausalBench \citep{chevalley2025large}, \textit{etc.}}]
            ]
        ] 
    ]
\end{forest}}
\caption{\label{taxonomy}
    Taxonomy of LLMs meet virtual cell
  }
\end{figure*}

\section{LLM Methods as Oracle for the Virtual Cell}
LLMs can be regarded as an computational \textit{Oracle} for the virtual cell, directly modeling the internal states and dynamics of cellular systems. In this mode, they operate on biological sequences, such as DNA, RNA, or single-cell transcriptomic profiles. The LLM itself serves as the predictive engine, learning representations of cellular components and interactions from raw data without relying on external tools. This approach emphasizes the model’s intrinsic capacity to encode and reason over biological information.

\subsection{Nucleotides}
DNA serves as the foundational blueprint of the cell, encoding not only protein-coding genes but also a vast regulatory landscape that governs when, where, and how genes are expressed \citep{IntegratedEncyclopediaDNA2012}. LLMs can act as powerful Oracles of regulatory mechanisms, enabling predictions of chromatin states, transcription factor binding, and the functional impact of genetic variants directly from nucleotide sequences \citep{tangEvaluatingRepresentationalPower2025}.

A key challenge in DNA modeling lies in capturing long-range dependencies: regulatory elements such as enhancers can influence gene expression from distances up to 100kb. Early models like \textbf{ExPecto} \citep{zhou2018deep} and \textbf{BPNet} \citep{long2023bpnet} addressed this using convolutional architectures, which excel at local pattern recognition but struggle with very long contexts. The advent of attention-based mechanisms, particularly the Transformer, overcame this limitation by enabling global context modeling \citep{dai2019transformer}. Combining CNNs with a Transformer backbone, \textbf{Enformer} \citep{avsec2021effective} enables the model to extend input sequences up to 200kb. More recent efforts have embraced pure Transformer encoder pretraining with masked language modeling (MLM). The \textbf{DNABERT} series \cite{ji2021dnabert,zhou2023dnabert} and the \textbf{Nucleotide Transformer (NT)} \citep{dalla2025nucleotide} are representative of this paradigm. In particular, NT scales up to 2.5 billion parameters. \textbf{HyenaDNA} \citep{nguyen2023hyenadna} replaces the standard self-attention mechanism with a novel Hyena operator and adopts autoregressive next-token prediction (NTP), enabling training and inference on sequences up to 1 million tokens. Besides, \textbf{Borzoi} \citep{linder2025predicting} predicts cell-type-specific RNA-seq coverage directly from DNA sequence.

RNA plays a diverse and active role in the cell, including catalyzing reactions, regulating gene expression, and serving as the template for protein synthesis \citep{wangBiochemistryRNAStructure2025}. To model these functions from sequence alone, \textbf{RNA-FM} \citep{shen2024accurate} is a transformer encoder–based model trained on 23.7 million non-coding RNA sequences. Building on a similar architecture, \textbf{RiNALMo} \citep{penic2025rinalmo} scales up to 650 million parameters. \textbf{RNAErnie} \citep{wang2024multi} employs motif-aware MLM during pretraining, enhancing its sensitivity to functional RNA elements. In contrast, \textbf{RNA-MSM} \citep{zhang2024multiple} uniquely leverages MSAs to capture evolutionary constraints. In addition, \textbf{SpliceBERT} \citep{chen2024self} is designed to predict splice sites and assess the impact of splicing-altering variants.

\subsection{Protein-protein Interactions}
Protein-protein interactions (PPIs) form the backbone of cellular signaling, complex assembly, and metabolic pathways \citep{nadaNewInsightsProtein2024}. One major PPI prediction method relies on evolutionary information from multiple sequence alignments (MSAs). \textbf{Alphafold-Multimer} \citep{evans2021protein} uses MSAs and pairwise features to predict high-accuracy 3D structures of protein complexes. Studies have shown that its predicted pDockQ metric can reliably distinguish PPIs \citep{bryant2022improved}. Similarly, \textbf{RoseTTAFold2-Lite} \citep{humphreys2024protein} and its variant \textbf{RoseTTAFold2-PPI} \citep{zhang2025predicting} offer fast and scalable alternatives for large-scale PPI screening. \textbf{AlphaMissense} \citep{cheng2023accurate}, on the other hand, assesses the functional impact of missense variants across the proteome, indirectly informing interaction stability.

However, MSA has its limitations regarding the high computational cost and reduced accuracy for sequences without close homologs, and these have motivated the development of protein language model-based (PLMs) approaches for predicting PPIs. \textbf{MINT} \citep{ullanat2025learning} is a scalable multimeric interaction transformer designed to model sets of interacting proteins, leveraging MLM. \textbf{SWING} \citep{siwek2025sliding} introduces a novel sliding window mechanism to capture the underlying grammar of peptide–protein interactions. At proteome scale, \textbf{ProteomeLM} \citep{malbranke2025proteomelm} employs a MLM framework to predict PPIs and gene essentiality across entire proteomes from multiple taxa.

\subsection{Multi-domain Molecules}
Comprehensive representation of multiple molecular types and their interactions can be a  key to capturing the complex dynamics and regulatory mechanisms underlying cell function.

\textbf{Evo} \citep{nguyen2024sequence} and its scaled successor \textbf{Evo2} \citep{brixi2025genome} are trained on trillions of nucleotides spanning all domains of life using a NTP approach. These models learn joint representations of DNA, RNA, and protein sequences, enabling downstream tasks such as variant effect prediction and genome design. Similarly, \textbf{LucaOne} \citep{he2025generalized} pretrains on nucleic acid and protein sequences from nearly 170,000 species using MLM.

On the other hand, state-of-the-art sequence-input structural prediction models have been extended to cover all types of biomolecules and their interactions, such as \textbf{RoseTTAFold-AA} \citep{krishna2024generalized}, \textbf{AlphaFold3}, \citep{abramson2024accurate} and \textbf{Chai-1} \citep{Chai-1-Technical-Report}. Notably, \textbf{Boltz-2} \citep{passaro2025boltz} is also capable of predicting both the likelihood and the strength of protein–small molecule binding, providing a quantitative assessment of molecular interactions.

\subsection{Single-omics}
Omics refers to large-scale molecular profiling technologies that capture the comprehensive molecular state of a cell \citep{micheelOmicsbasedClinicalDiscovery2012}. These data collectively reflect a cell’s status \citep{hasinMultiomicsApproachesDisease2017a}. The fundamental data structure for single-cell omics is normally a cell-by-gene expression matrix $\mathbf{X} \in \mathbb{R}^{N \times G}$, where $N$ denotes the number of cells and $G$ the number of genes profiled. 

Among single-cell omics methods, single-cell RNA sequencing (scRNA-seq) has become the dominant data source for foundational LLMs in cell modeling \citep{rizviScalingLargeLanguage2025}. This prevalence stems from key advantages such as functional relevance, as the transcriptome reflects the cell’s active state, and data abundance. Given the inherent challenges of omics data, including noise and batch effects,  
\textbf{xTrimoGene} \citep{gong2023xtrimogene} and \textbf{scFoundation} \citep{hao2024large} employ a masked autoencoder \citep{he2022masked} (MAE)-like architecture, where a subset of input is masked during training and the model learns to reconstruct them from the observed context. Similar to the MLM approach in \textbf{scBERT} \citep{yang2022scbert}, \textbf{Geneformer}  \citep{theodoris2023transfer} scales its training set to 30 million cells, while \textbf{AIDO.Cell} \citep{ho2024scaling} further scales to 50 million cells and expands the model size up to 650 million parameters. In contrast, \textbf{CellFM} \citep{zeng2025cellfm} explores architectural innovation by replacing the standard Transformer with a modified ERetNet backbone. Meanwhile, \textbf{tGPT} \cite{shen2023generative} adopts an NTP objective with an autoregressive Transformer decoder.

Beyond architectural choices, incorporating biological priors into the modeling process has proven effective for task-specific enhancement \citep{liuAdvancingBioinformaticsLarge2025}. For instance, \textbf{GeneCompass} \citep{yang2024genecompass} integrates external biological meta data to better capture gene regulatory mechanisms. To improve cross-species generalization, \textbf{UCE} \citep{rosen2024toward} and \textbf{scPRINT} \citep{kalfon2025scprint} augment gene tokens with embeddings of their most common protein products derived from PLM ESM-2 \citep{evans2021protein}. \textbf{TranscriptFormer} \citep{pearce2025cross} extends this idea further by adopting an NTP-based autoregressive framework trained on an unprecedented scale of 112 million cells from 12 species. Besides, \textbf{STATE} \citep{adduri2025predicting} is specifically designed for perturbation response prediction: it is pretrained on nearly 170 million unperturbed cells and fine-tuned using perturbational data from over 100 million cells across 70 species.

Epigenomic modification regulate gene expression without altering DNA sequence, acting as a critical layer of cellular memory and identity. \textbf{scBasset} \citep{yuan2022scbasset} predicts chromatin accessibility directly from DNA sequence, using a convolutional architecture. More recently, \textbf{EpiGePT} \citep{gao2024epigept} integrating sequence, chromatin, and genome into a transformer encoder-based foundation model, enabling context-aware prediction of epigenomic states across cell types.

\subsection{Multi-omics}
A central challenge in modeling the virtual cell is that no single omics fully captures cellular state: chromatin accessibility defines regulatory potential, gene expression reflects functional output, and protein abundance mediates phenotypic effects. Multi-omics integration therefore offers a promising solution to capture the full complexity of cellular behavior \citep{baysoyTechnologicalLandscapeApplications2023}.

\textbf{scGPT} \citep{cui2024scgpt} introduced a GPT-style autoregressive architecture that tokenizes diverse omics data into a shared vocabulary, enabling unified modeling of multi-omic profiles through language modeling objectives. Its spatial extension, \textbf{scGPT-spatial} \citep{wang2025scgpt}, further incorporates tissue coordinates as additional tokens, allowing joint modeling of cellular profiles and spatial context. \textbf{spaLLM} \citep{li2025spallm}, which is also built upon scGPT, integrates graph neural networks (GNNs) to explicitly model cell–cell neighborhood relationships in spatial transcriptomics data. Similarly, \textbf{GLUE} \citep{cao2022multi} employs a graph-involved variational autoencoder to align scRNA-seq, scATAC-seq, and snmC-seq into a common latent space. In contrast, \textbf{GET} \citep{fu2025foundation} adopts a Enformer-like hybrid CNN–transformer architecture for processing scATAC-seq and scRNA-seq. Built upon a similar architecture, \textbf{EpiBERT} \citep{javed2025multi} adopts a masked modeling pretraining strategy while integrating DNA sequences and scATAC-seq data. Most ambitiously, \textbf{PertFormer} \citep{yang2024multiomic} scales to a 3B model pretrained on 9 distinct single-cell omics, capable for zero shot prediction on diverse downstream tasks.

Complementing data integration efforts, multi-omic translation seeks to infer or reconstruct missing omic modalities from available data, enabling more complete cellular representations. \textbf{scPER2P} \citep{wang2024scper2p} employs a transformer decoder architecture to translate scRNA-seq inputs into corresponding proteome profiles. Similarly, \textbf{scTEL} \citep{chen2025joint} is specifically designed to map scRNA-seq profiles to their matched CITE-seq measurements at single-cell resolution.

\subsection{Multi-modal}
Beyond cellular data, recent studies have begun to leverage the general language understanding capabilities of LLMs, incorporating scientific text as an additional modality to ground cellular predictions and enhance task generalization. 

\textbf{CellWhisperer} \citep{schaefer2024multimodal}  adopts a CLIP-like contrastive learning framework, aligning latent representations of scRNA-seq profiles and textual description in a shared space.
\textbf{C2S (Cell2Sentence) } \citep{levine2024cell2sentence} takes a different approach, it using value binning approach to tokenize genes and mapping them to a fixed vocabulary. This enables direct fine-tuning of GPT-2, allowing the text LLM to process scRNA-seq data. 

\textbf{scMMGPT} \citep{shi2025language} performs text–gene alignment, which is analogous to BLIP-2’s text–image alignment framework \citep{li2023blip}. Unlike BLIP-2, it integrates a single-cell LLM and a general-purpose text LLM, which are linked through bidirectional cross-attention between cell and text latent. This architecture enables bidirectional translation between cellular and textual modalities.
Similarly, \textbf{InstructCell} \citep{fang2025multi}  leverages a Q-Former module to extract representations from scRNA-seq data, which are then injected as soft prompts into a T5-base LM. On the other hand, \textbf{ChatNT} \citep{richard2024chatnt} unifies DNA, RNA, protein sequences, and natural language in a single system. It combines NT v2 as a molecular encoder with Vicuna-7B LM as its textual backbone. 

Emerging systems move beyond passive data fusion, aiming to enable scientific reasoning. Reinforcement learning, which has recently proven effective in improving the reasoning ability of general LLMs \citep{team2025qwq, team2025kimi}, offers a potential pathway to endow virtual cell models with more autonomous discovery capabilities.
\textbf{C2S-Scale} \citep{rizvi2025scaling} employs GRPO \citep{guo2025deepseek} to align scRNA-seq representations with natural language understanding and deductive reasoning. At inference time, chain-of-thought (CoT) prompting has proven highly effective for eliciting step-by-step reasoning from LLMs \citep{wei2022chain}. \textbf{CellReasoner} \cite{cao2025cellreasoner} leverages this by distilling CoT rationales generated by the DeepSeek-R1-671B into supervised fine-tuning signals, thereby endowing its 7B architecture with reasoning abilities. Building on both strategies,  \textbf{rBio1} \citep{istrate2025rbio1} integrates GRPO-style RL post-training with test-time CoT, achieving advanced performance in tasks such as perturbation effect prediction.

\section{LLM Methods as Agent for the Virtual Cell}
LLMs can also function as intelligent \textit{agents} for the virtual cell, orchestrating external tools, databases, and simulation environments to accomplish more complex scientific research tasks that go beyond traditional modeling, generative, and predictive functions \citep{harrerGenerativeAIAgents2024}. Unlike foundation models that passively generate outputs from learned representations, LLM agents actively plan, reason, and act within an adaptive and goal-driven framework \citep{huang2024understanding}.

\subsection{Architecture} 
From an architectural standpoint, virtual cell LLM agents can be divided into single-agent and multi-agent frameworks. The choice between them often depends on the complexity of task, computational cost, and the desired level of interpretability.

In single-agent systems, a single LLM operates as a unified intelligence, managing the entire workflow through internal reasoning or dynamic prompting \citep{huang2025biomni}. Such designs often rely on structured system prompts or internal role-switching mechanisms to simulate modularity without invoking separate models \citep{xin2024bioinformatics}. 

In contrast, multi-agent systems distribute responsibilities across multiple specialized LLMs, each serving as an autonomous agent (e.g., planner, analyzer, or executor) that collaborates through dialogue or shared memory \citep{tang2025scagents}. This design facilitates scalability, transparency, and division of labor in complex cellular modeling pipelines \citep{ yiyao2025omicsnavigator, wang2025accelerating}. 

\subsection{Literature \& Knowledge}
To ensure factual accuracy and biological validity, LLM-based agents increasingly interface with scientific literature and structured data repositories, which not only provide verified information but also enhance their reasoning capabilities through access to authoritative knowledge sources \citep{zhangEvolvingRoleLarge2025}. A particularly effective strategy is Retrieval-Augmented Generation (RAG) \citep{lewis2020retrieval}, which enhances model responses by retrieving and incorporating relevant information at inference time, thereby improving factual accuracy and reducing hallucination. For instance, \textbf{BioRAG} \citep{wang2024biorag} indexes over 22 million scientific articles to deliver factually grounded answers to complex biological questions. Similarly, \textbf{GENEVIC} \citep{nath2024genevic} leverages RAG to provide interactive access to domain-specific knowledge bases such as PubMed. Beyond literature, \textbf{CompBioAgent} \citep{zhang2025compbioagent} is an LLM-based agent that focuses on scRNA-seq resources, allowing users to query gene expression patterns via intuitive natural language interfaces. 

LLM-based agents are also being deployed for data curation and management for virutal cell research. \textbf{SRAgent} \citep{youngblut2025scbasecount}, for instance, autonomously harvests and processes scRNA-seq data to construct \textit{scBaseCount}, a continuously expanding database.

\subsection{Experimental Design}
LLM-based agents are increasingly employed to support the design of virtual cell experiments, transforming high-level biological questions into actionable experimental plans \citep{renScientificIntelligenceSurvey2025}.  \textbf{SpatialAgent} \citep{wang2025spatialagent} interprets spatial transcriptomics data to propose novel mechanistic hypotheses about tissue organization and cellular interactions. \textbf{PROTEUS} \citep{ding2024automating} enables discovering from proteomics datasets and generating novel, data-driven biological hypotheses without manual intervention.

In parallel, LLM agents also excel at process instruction, translating abstract research goals into concrete, step-by-step experimental protocols. \textbf{CRISPR-GPT} \citep{huang2024crispr} is an LLM-powered agent designed for CRISPR-based gene-editing workflows, which automatically decompose the entire design process and leverages domain knowledge to narrow down options to a focused set of high-quality candidates. \textbf{PerTurboAgent} \citep{hao2025perturboagent} is capable of guiding the experiments in functional genomics by planning iterative Perturb-Seq experiments, intelligently selecting optimal gene panels for successive rounds of perturbation to maximize biological insight. Meanwhile, \textbf{BioResearcher} \citep{luo2025intention} employs RAG framework to ground its reasoning in the most relevant scientific literatures, and converting high-level research intentions into executable experimental pipelines.

\subsection{Computational Workflow Automation}
Beyond experimental design, LLM agents can play an instrumental role in automating complex computational workflows in virtual cell research. For instance, \textbf{CellAgent} \citep{xiao2024cellagent} can perform end-to-end interpretation of single-cell RNA-seq and spatial transcriptomics data through natural language interaction. 
Similarly, \textbf{AutoBA} \citep{zhou2024ai} can autonomously construct adaptive multi-omic analysis pipelines with minimal user input, demonstrating robust performance across diverse datasets and analytical contexts. 

At a more integrative level, agents can go beyond analysis to actively build and operate virtual cell models. \textbf{CellForge} \citep{tang2025cellforge} is designed to autonomously construct predictive computational models of cellular behavior directly from raw omics data and high-level task descriptions, enabling applications in tasks like perturbation prediction. \textbf{BioMaster} \citep{su2025biomaster} enhances long-horizon workflow execution by integrating RAG and optimizing agent coordination for extended pipelines.

\subsection{Full-stack Research}
At the frontier of LLM-based agents for cellular research, full-stack research agents aim to automate the entire scientific workflow, from question formulation to discovery. \textbf{CellVoyager} \citep{alber2025cellvoyager} operates in general computational biology settings, autonomously analyzing diverse omics data to produce novel insights—bypassing fixed task templates by using iterative self-querying and tool-augmented reasoning to explore data-driven hypotheses. \textbf{BioDiscoveryAgent} \citep{roohani2024biodiscoveryagent} focuses on functional genomics and disease mechanism discovery; it implements full-stack research by iteratively proposing genetic perturbations, simulating their outcomes using in silico models, evaluating results, and refining hypotheses in a closed loop. \textbf{OmniCellAgent} \citep{huang2025omnicellagent} targets precision medicine applications, where it translates questions into multi-omic analyses and delivers interpretable reports, effectively managing the entire research lifecycle.

\subsection{Optimization}
To enhance the reliability, accuracy, and adaptability of LLM agents in virtual cell applications, recent work has introduced sophisticated optimization strategies that operate at multiple stages of the agent lifecycle. An effective approach is post-training via reinforcement learning. For instance, \textbf{Biomni-R0} \citep{biomnir0} employs multi-turn reinforcement learning \cite{guo2025deepseek} across a diverse suite of biomedical tasks, yielding agentic LLMs that significantly outperform their base models. 

Beyond post-training, agents can also self-refine during inference by iteratively verifying and correcting their outputs. \textbf{GeneAgent} \citep{wang2025geneagent} implements a self-verification mechanism that cross-references authoritative biological databases in real time during gene-set analysis, drastically reducing hallucinations and improving biological fidelity. Similarly, \textbf{TransAgent} \citep{zhang2025transagent} dynamically refines its interpretation of transcriptional regulatory networks by integrating feedback from multi-omics data streams. \textbf{PhenoGraph} \citep{niyakan2025phenograph} grounding its spatial phenotype discovery in structured knowledge graphs, ensuring hypotheses are both data-driven and biologically plausible. Additionally, \textbf{BioAgents} \citep{mehandru2025bioagents} adopts an \textit{agent-as-judge} \cite{zhuge2024agent} method, where specialized evaluator agents perform self-assessment of outputs to enhance overall reliability.

Self-evolution agents aim to continuously accumulate knowledge and improve their reasoning strategies over time. \textbf{OriGene} \citep{zhang2025origene} achieves this through a dual-loop system: it uses a ReAct-style \citep{yao2023react} iterative reflection-and-replanning process for task execution, while also maintaining a library of reasoning templates involving human experts that evolves with expert feedback. Similarly, \textbf{STELLA} \citep{jin2025stella} implements a self-evolving architecture by iteratively updating its Template Library for reasoning patterns, and expanding its accessible Tool Ocean, a dynamic inventory of computational tools.

\section*{Conclusion and Future Work}
This paper presents a comprehensive survey of LLMs for the virtual cell. We first introduced various virtual cell tasks and their evaluation protocols. We then categorized existing methods into two major paradigms: LLM as Oracle and LLM as Agent, and highlight their respective architectures and applications. These works represent significant advancements in virtual cell research. However, important challenges and opportunities remain for the future:

\textbf{Scalability} For LLM Oracles, scalability demands unifying multiple modalities, spanning molecular-level and omics-level sequences into coherent, joint representations. It also requires adopting efficient architectures capable of handling ultra-long cellular contexts. For LLM Agents, scalability hinges on long-term memory mechanisms that maintain coherent reasoning and contextual awareness over extended experimental workflows, enabling consistent planning across dozens of tool invocations and iterative hypothesis refinement.

\textbf{Generlizability \& Benchmarking}
For LLM Oracles, generalization to unseen cell types remains a significant challenge \citep{ahlmann2025deep}. Addressing this requires not only advances in training strategies and model architectures but also the development of more rigorous and biologically meaningful benchmarks.
Similarly, LLM Agents currently lack systematic and fair evaluation frameworks. The absence of standardized tasks, environments, and metrics hinders our understanding of their strengths and weaknesses.

\textbf{Reliability \& Interpretability}
LLM Oracles require stability to ensure reliable, reproducible simulations, with uncertainty estimation and interpretability to quantify prediction confidence. Meanwhile,  LLM Agents need stability for consistent behavior, using uncertainty and interpretability to make decisions understandable and verifiable.

\section*{Limitions}
This survey centers on the intersection between LLMs and virtual cell research. We recognize that the study of cellular imaging represents a rich and expansive field. However, given its considerable breadth, we do not cover this area extensively in the present work. Our focus remains on LLMs applied to tasks that are primarily centered around virtual cells. In future work, we may broaden our scope to provide a more complete of those domains.

\bibliography{custom}

\appendix

\section{Appendix}
\label{sec:appendix}
\subsection{Tokenization methods for Biological Sequences}
\subsubsection{DNA \& RNA}
DNA and RNA sequences can be naturally tokenized using k-mers \citep{ji2021dnabert} or subword units such as Byte Pair Encoding borrowed from natural language processing \citep{zhou2023dnabert}, enabling direct application of LLMs.
\subsubsection{Protein}
Protein sequences are naturally represented as strings of single-letter amino acid codes, where each character corresponds to one residue in the polypeptide chain \citep{lin2023evolutionary}. 
\subsubsection{Single-omics}
To adapt the continuous, high-dimensional, and sparse matrix of omics data for language modeling, recent LLMs have developed several principled tokenization strategies:

(1) \textbf{Top-$k$ gene selection}: Only the $k$ most highly expressed genes per cell are retained, and treating each gene symbol as a token \citep{theodoris2023transfer,shen2023generative}.  

(2) \textbf{Value binning}: Continuous expression values are discretized into bins, and each pair is mapped to a unique token \citep{yang2022scbert}.  

(3) \textbf{Projection-based embedding}: The entire expression vector is projected through a learnable linear layer into a dense embedding space, bypassing explicit tokenization \citep{lopez2018deep,gong2023xtrimogene}.

\subsection{Evaluation Metrics for Major Tasks in AI Virtual Cell}

\subsubsection{Cellular Representation}
\begin{itemize}
    \item \textbf{Normalized Mutual Information (NMI):} Measures the similarity between predicted clusters and true labels for cell clustering, normalized to [0,1]. Higher values indicate better clustering.
    \begin{equation}
        \text{NMI}(U, V) = \frac{2 \cdot I(U;V)}{H(U) + H(V)}
    \end{equation}
    where $I(U;V)$ is the mutual information between cluster assignment $U$ and ground truth $V$, and $H(\cdot)$ is the entropy.
    
    \item \textbf{Accuracy (ACC):} Fraction of correctly classified samples for cell type classification.
    \begin{equation}
        \text{ACC} = \frac{\text{Number of correctly classified samples}}{\text{Total number of samples}}
    \end{equation}
    
    \item \textbf{Precision:} Fraction of true positive predictions among all positive predictions for cell type classification.
    \begin{equation}
        \text{Precision} = \frac{\text{TP}}{\text{TP + FP}}
    \end{equation}
    
    \item \textbf{Recall:} Fraction of true positive predictions among all actual positives.
    \begin{equation}
        \text{Recall} = \frac{\text{TP}}{\text{TP + FN}}
    \end{equation}
    
    \item \textbf{Macro F1:} Harmonic mean of precision and recall computed per class and averaged for cell type classification.
    \begin{equation}
        \text{Macro F1} = \frac{1}{C} \sum_{i=1}^{C} \frac{2 \cdot \text{Precision}_i \cdot \text{Recall}_i}{\text{Precision}_i + \text{Recall}_i}
    \end{equation}
    where $C$ is the number of classes.
\end{itemize}

\subsubsection{Perturbation Prediction}

\begin{itemize}
    \item \textbf{Mean Squared Error (MSE):} Measures the average squared difference between predicted and true values.
    \begin{equation}
        \text{MSE} = \frac{1}{N} \sum_{i=1}^{N} (y_i - \hat{y}_i)^2
    \end{equation}

    \item \textbf{Root Mean Squared Error (RMSE):} Square root of MSE, representing error in the same units as the target.
    \begin{equation}
        \text{RMSE} = \sqrt{\text{MSE}}
    \end{equation}

    \item \textbf{Recall:} Fraction of true positives correctly identified.
    \begin{equation}
        \text{Recall} = \frac{\text{TP}}{\text{TP + FN}}
    \end{equation}

    \item \textbf{False Discovery Proportion (FDP):} Fraction of false positives among all positive predictions.
    \begin{equation}
        \text{FDP} = \frac{\text{FP}}{\text{TP + FP}}
    \end{equation}

    \item \textbf{Pearson Correlation:} Measures linear correlation between predicted and true values.
    \begin{equation}
        r = \frac{\sum_{i=1}^{N} (y_i - \bar{y})(\hat{y}_i - \bar{\hat{y}})}{\sqrt{\sum_{i=1}^{N} (y_i - \bar{y})^2} \sqrt{\sum_{i=1}^{N} (\hat{y}_i - \bar{\hat{y}})^2}}
    \end{equation}

    \item \textbf{Spearman Correlation:} Measures rank correlation between predicted and true values.
    \begin{equation}
        \rho = 1 - \frac{6 \sum_{i=1}^{N} d_i^2}{N(N^2 - 1)}
    \end{equation}
    where $d_i$ is the difference between ranks of $y_i$ and $\hat{y}_i$.
\end{itemize}
\subsubsection{Gene Functions \& Regulations Prediction}
\begin{itemize}
    \item \textbf{Area Under the Precision-Recall Curve (AUPRC):} Measures overall prediction quality, especially for imbalanced datasets. Higher values indicate better precision-recall trade-off.
    \begin{equation}
        \text{AUPRC} = \int_0^1 \text{Precision}(\text{Recall}) \, d\text{Recall}
    \end{equation}
    
    \item \textbf{Early Precision Ratio (EPR):} Evaluates precision among the top-ranked predictions, emphasizing early retrieval of correct hits.
    \begin{equation}
        \text{EPR@k} = \frac{\text{\# true positives in top-}k}{k}
    \end{equation}

    \item \textbf{Enrichment Score (ES):} Measures whether genes of interest are overrepresented at the top of a ranked list. Following the \textit{prerank} methodology:
    \begin{itemize}
        \item \textbf{Target-Hub:} Sum all edge scores of the adjacency matrix row-wise:
        \begin{equation}
            ES_{\text{row}} = \sum_{j} M_{ij} \quad \forall i \in \text{target genes}
        \end{equation}
        \item \textbf{Regulator-Hub:} Sum all edge scores column-wise:
        \begin{equation}
            ES_{\text{col}} = \sum_{i} M_{ij} \quad \forall j \in \text{regulator genes}
        \end{equation}
        \item \textbf{Network Centrality:} Compute eigenvector centrality of nodes using NetworkX, with prerank background comprising all genes:
        \begin{equation}
            ES_{\text{centrality}} = \text{eig\_centrality}(G)
        \end{equation}
    \end{itemize}
\end{itemize}

\subsection{Datasets}

\subsubsection{Pre-training Datasets}
\textbf{CELLxGENE} \citep{czi2025cz}, maintained by the Chan Zuckerberg Initiative (CZI), is one of the world’s largest standardized portals for scRNA-seq data. It offers over 120 millions of curated and standardized data, and allows flexible data slicing based on metadata (e.g., tissue, donor, condition).

\textbf{NCBI GEO} (Gene Expression Omnibus) \citep{clough2024ncbi} is a public repository for high-throughput functional genomics data, including over 8 millions of samples. It provides diverse gene expression profiles across conditions, tissues, and disease contexts.

\textbf{ENA} (European Nucleotide Archive) \citep{leinonen2010european} is a comprehensive repository of raw sequencing reads, alignments, and assemblies for DNA and RNA experiments worldwide. It provides base-level sequence information that allows models to learn genomics, transcript variants, and mutation patterns.

\textbf{ImmPort} \citep{bhattacharya2014immport} contains raw and processed data from more than 170 clinical trials, mechanistic studies, and molecular assays, offering immunology-focused datasets linking molecular features to clinical and cellular phenotypes.

\textbf{scBaseCount} \citep{youngblut2025scbasecount} is a single-cell RNA-seq database. It integrates over 300 million cells across 26 species and 72 tissues, automatically processed and updated by SRAgent.

\textbf{GeneOntology} (GO) \citep{gene2004gene} provides a unified, structured vocabulary describing gene functions through three ontologies: molecular function, cellular component, and biological process. It stands for a foundational resource for biological annotation.

\textbf{Protein Data Bank} \citep{sussman1998protein} is one of the largest repositories of macromolecular structures and their interactions, providing a rich resource for training models that learn molecular-level representations and interactions.

\subsubsection{Cellular Representation Benchmarks}

\textbf{Segerstolpe dataset} \citep{abdelaal2019comparison} includes scRNA-seq data from 2209 (2133 after processed) pancreatic cells across 10 distinct cell populations, derived from both healthy donors and individuals with type 2 diabetes, making it a standard for evaluating cell type classification in a disease context.

\textbf{Zheng68K} \citep{hou2020systematic} is collected from human peripheral blood mononuclear cells (PBMC), is also a benchmark for cell type classification. This dataset consists of scRNA-seq profiles from approximately 68,000 PBMCs.

\textbf{Tabula Sapiens V2} \citep{quake2011tabula} contains over 0.5 million cells with 27 tissues sampled from both male and female donors. It allows to evaluate model performance for cell clustering, classification, and metadata prediction based on gene expression counts. 

\textbf{Spermatogenesis} \citep{murat2023molecular} provides a cross-species, single-nucleus transcriptomic resource focused on the mammalian testis. It able to evaluate the performance for cell type prediction across species based on gene expression counts.

\subsubsection{Perturbation Prediction Benchmarks}

\textbf{Adamson dataset} \citep{adamson2016multiplexed} is a Perturb-seq dataset that applies CRISPR interference (CRISPRi) to dissect the mammalian unfolded protein response (UPR). It provides single-cell transcriptional profiles in response to a large number of single-gene perturbations, serving as a primary benchmark for single-perturbation effect prediction.

\textbf{Norman dataset} \citep{norman2019exploring} extends beyond single perturbations to include combinatorial (dual-gene) knockouts. This feature makes it a key resource for evaluating a model's capacity to capture non-linear, epistatic interactions between genes.

\textbf{Systema} \citep{vinas2025systema} is a more recent benchmark for perturbation prediction. It is explicitly designed to assess whether models predict true biological signal or merely capture systematic, non-biological variation inherent in perturbation.

\subsubsection{Gene Functions \& Regulations Prediction Benchmarks}
\textbf{BEELINE} \citep{akers2021gene} stands for a 'de facto' standard benchmark for GRN inference. It provides a suite of both simulated and real scRNA-seq datasets, each paired with a high-confidence "ground truth" regulatory network.

\textbf{geneRNIB} \citep{nourisa2025genernib} is built on three core principles: context-specific evaluation, continuous integration of new methods and data, and holistic assessment, aiming to provide a more dynamic and comprehensive evaluation of GRN inference.

\textbf{CausalBench} \citep{chevalley2025large} leveraging large-scale, real-world single-cell perturbation data as its foundation for evaluation. It provides a more biologically grounded and causal assessment of network inference methods compared to benchmarks that rely primarily on observational or simulated data.

Besides, \textbf{scEval} \citep{liu2023evaluating} is comprehensive evaluation platform for single-cell foundation models. scEval provides a holistic assessment by evaluating model performance across eight diverse downstream tasks, including cell annotation, perturbation prediction, and GRN inference.

\end{document}